\def\year{2020}\relax
\documentclass[letterpaper]{article} 
\usepackage{aaai20}  
\usepackage{times}  
\usepackage{helvet} 
\usepackage{courier}  
\usepackage[hyphens]{url}  
\usepackage{graphicx} 
\usepackage{graphicx}  
\usepackage{epsfig}
\usepackage{graphicx}
\usepackage{subfigure}
\usepackage{overpic}
\usepackage{amsfonts}
\usepackage{booktabs}
\usepackage{bm}
\usepackage{amsmath}
\usepackage{bbm}
\usepackage{url}

\newcommand{\figref}[1]{Figure \ref{#1}}
\newcommand{\tabref}[1]{Table \ref{#1}}
\newcommand{\equref}[1]{Eq.\ref{#1}}

\title{DMRM: A Dual-channel Multi-hop Reasoning Model for Visual Dialog}

\author{
Feilong Chen,\textsuperscript{\rm 1,2,3,4}\thanks{This work was done when Feilong Chen was interning at
Pattern Recognition Center, WeChat AI, Tencent Inc., China} Fandong Meng,\textsuperscript{\rm 2} Jiaming Xu,\textsuperscript{\rm 1,3}\thanks{Corresponding author.} Peng Li,\textsuperscript{\rm 2} Bo Xu,\textsuperscript{\rm 1,3,4,5} Jie Zhou\textsuperscript{\rm 2} \\ 
\textsuperscript{\rm 1}Institute of Automation, Chinese Academy of Sciences (CASIA), Beijing, China.\\
\textsuperscript{\rm 2}Pattern Recognition Center, WeChat AI, Tencent Inc., China\\
\textsuperscript{\rm 3}Research Center for Brain-inspired Intelligence, CASIA\\
\textsuperscript{\rm 4}University of Chinese Academy of Sciences\\
\textsuperscript{\rm 5}Center for Excellence in Brain Science and Intelligence Technology, CAS. China\\
\{chenfeilong2018, jiaming.xu, xubo\}@ia.ac.cn\\
\{fandongmeng, patrickpli, withtomzhou\}@tencent.com\\
}

\begin{document}

\maketitle

\begin{abstract}
Visual Dialog is a vision-language task that requires an AI agent to engage in a conversation with humans grounded in an image. It remains a challenging task since it requires the agent to fully understand a given question before making an appropriate response not only from the textual dialog history, but also from the visually-grounded information. While previous models typically leverage single-hop reasoning or single-channel reasoning to deal with this complex multimodal reasoning task, which is intuitively insufficient. In this paper, we thus propose a novel and more powerful Dual-channel Multi-hop Reasoning Model for Visual Dialog, named DMRM. DMRM synchronously captures information from the dialog history and the image to enrich the semantic representation of the question by exploiting dual-channel reasoning. Specifically, DMRM maintains a dual channel to obtain the question- and history-aware image features and the question- and image-aware dialog history features by a mulit-hop reasoning process in each channel. Additionally, we also design an effective multimodal attention to further enhance the decoder to generate more accurate responses. Experimental results on the VisDial v0.9 and v1.0 datasets demonstrate that the proposed model is effective and outperforms compared models by a significant margin.

\end{abstract}

\section{Introduction}
With the rapid development of both computer vision and natural language processing, visual-language tasks such as image caption~\cite{xu2015show,Anderson2016SPICE,anderson2018bottom} and visual question answering~\cite{ren2015exploring,gao2015you,lu2016hierarchical,anderson2018bottom} have attracted increasing attention in recent years. Although these tasks have inspired tremendous efforts on integrating vision and language to develop smarter AI, they are mostly {\em single-round} while human conversations are generally multi-round. Therefore, the Visual Dialog task is proposed to encourage research on multi-round visually-grounded dialog by~\citeauthor{das2017visual}~\shortcite{das2017visual}.


In Visual Dialog, an agent is required to answer a question given the dialog history and the visual context. In order to make an appropriate response, it is necessary for the agent to gain a proper understanding of the question, which requires it to exploit the textual dialog history and the visual context. To this end, some studies~\cite{das2017visual,lu2017best} design models to obtain features from both modalities. 
\citeauthor{das2017visual}~\shortcite{das2017visual} propose Late Fusion (LF), which directly concatenates individual representations of the question, the dialog history, and the image, and then generates a new joint representation by a linear transformation on them. 
\citeauthor{lu2017best}~\shortcite{lu2017best} design a history-conditioned attention image encoder to generate the representation of the question, the question-aware dialog history and the history-conditioned image features, and then concatenate them to be joint representations. 

Nevertheless, the approaches mentioned above are of single-hop approaches, which show the limited ability of reasoning and neglect latent information of the interactions among the question, the dialog history and the image.
For better solutions, researchers~\cite{das2017visual,wu2018you,niu2019recursive,kang2019dual} investigate multi-hop reasoning approaches~\cite{hudson2018compositional,hu2018explainable} to conduct interactions among modalities. For example, \citeauthor{wu2018you}~\shortcite{wu2018you} provide a sequential co-attention encoder, which firstly obtains question-aware image features, secondly extracts history features by co-attention mechanism with the question features and the extracted image features, thirdly gets the attended question features by the extracted history features and image features, and finally joints the three attended features and send them to the decoder. \citeauthor{niu2019recursive}~\shortcite{niu2019recursive} propose a recursive visual attention model, which recursively reviews the dialog history to find the reference of the question, and then extracts the image features by the attention model with the extracted history context and the question. 
These approaches are of single-channel approaches, which firstly use the question to find reference from the dialog history and then extract the image context from both the question and the history context. However, humans usually deal with a visually-grounded, multi-turn dialog by simultaneously comprehending the two aspects of information, namely both the textual dialog history and the visual context. That is to say, the question can find reference first from the image and then form the dialog history to enrich the question representation, and vice versa. 

Dual-channel reasoning, i.e., acquiring information from the dialog history and the image simultaneously, is beneficial for gaining an original understanding of the question from the dialog history and the image. Meanwhile, multi-hop reasoning, i.e., reasoning among the question, the dialog history and the image, is conducive to utilizing abundant latent information among the three inputs. Therefore, in this paper, we propose a Dual-channel Multi-hop Reasoning Model for Visual Dialog, named DMRM. DMRM synchronously captures information from the dialog history and the image to enrich the semantic representation of the question by exploiting dual-channel reasoning, which is composed of a Track Module and a Locate Module.
Track Module aims to enrich the representation of the question from the visual information while Locate Module aims to reach the same goal from the textual dialog history.
Specifically, DMRM maintains dual channels to obtain the question- and history-aware image features and the question- and image-aware dialog history features by a mulit-hop reasoning process in each channel. In addition, we design an effective multimodal attention to further enhance the decoder to generate more accurate responses.

We validate the DMRM model on large-scale datasets: VisDial v0.9 and v1.0~\cite{das2017visual}. DMRM achieves the state-of-the-art results on some metrics compared to other methods. We also conduct ablation studies to demonstrate the effectiveness of our proposed components. Furthermore, we conduct the human evaluation to indicate the effectiveness of our model in inferring answers.

Our main contributions are threefold:
\begin{itemize}
  \item We propose a dual-channel multi-hop reasoning model to deal with this complex multimodal reasoning task which enriches the semantic representation of the question, and thus the agent can make an appropriate response.
  \item We are the first to apply multimodal attention to the decoder for visual dialog and demonstrate the necessity and effectiveness of this attention mechanism for the decoding of visual dialog.
  \item We evaluate our method on two large-scale datasets and conduct ablation studies, human evaluation. Experimental results on VisDial v0.9 and v1.0 demonstrate that the proposed model achieves the state-of-the-art results on some metrics\footnote{Code is available at https://github.com/phellonchen/DMRM.}.
\end{itemize}

\begin{figure*}[t!]
\centering
  \begin{overpic}[width=\textwidth]{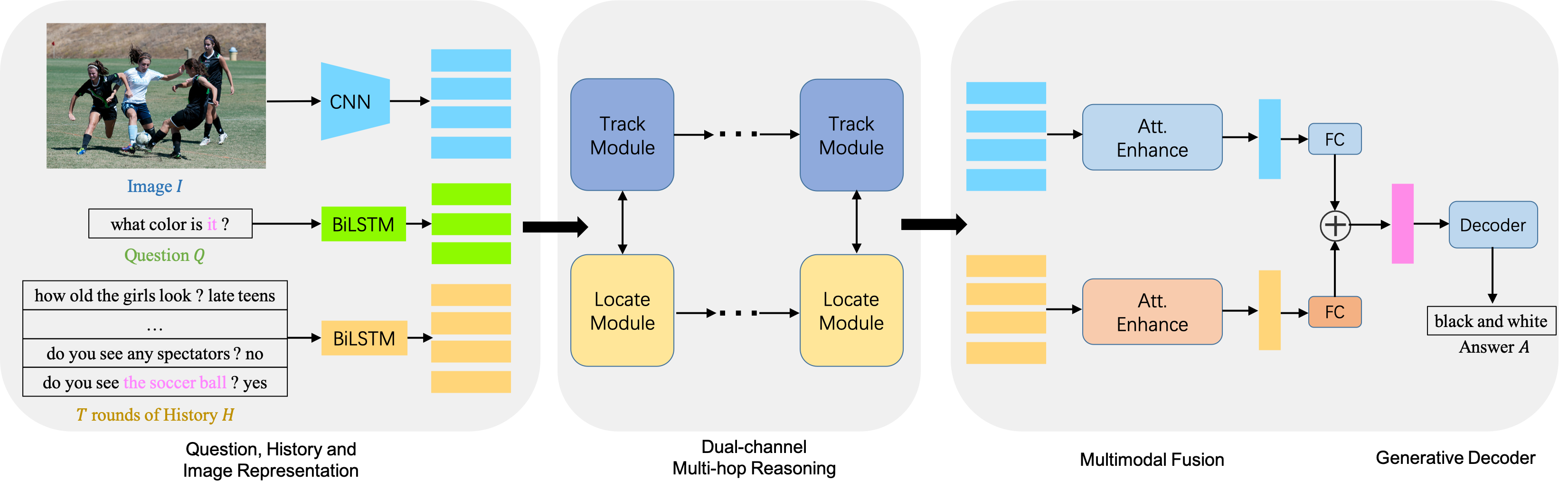}
   \end{overpic}
   \caption{The framework of the DMRM model. DMRM synchronously captures information from the dialog history and the image to enrich the semantic representation of the question by exploiting dual-channel reasoning, which is composed of Track Module and Locate Module. Track Module aims to make a fully understanding of the question from the aspect of the image. Locate Module aims to make a fully understanding of the question from the aspect of the dialog history. Finally, the outputs of the dual-channel reasoning are sent to the decoder after att-enhanece and multimodal fusion operation. }\label{fig:visualmodel}
\end{figure*}

\section{Our Approach}

In this section, we formally describe the visual dialog task and our proposed method, Dual-channel Multi-hop Reasoning Model (DMRM). According to \citeauthor{das2017visual}\shortcite{das2017visual}, inputs of a visual dialog agent consist of an image $I$, a caption $C$ describing the image, a dialog history (question-answer pairs) till round $t-1$: $H=(\underbrace{C}_{H_0}, \underbrace{(Q_1, A_1)}_{H_1}, \dots, \underbrace{(Q_{t-1}, A_{t-1})}_{H_{t-1}})$ and the current question $Q_t$ at round $t$. The goal of the visual dialog agent is to generate a response $A_t$ to the question $Q_t$.

Given the problem setup, DMRM for visual dialog consists of four components: (1) Input Representation, where the representations of the image and the textual information are generated for reasoning; (2) Dual-channel Multi-hop Reasoning, where our reasoning is applied to encode input representations; (3) Multimal Fusion, where we fuse the multimodal information; (4) Generative Decoder, where we use our multimodal attention decoder to generate the response. Specifically, we use Track Module and Locate Module to implement our dual-channel multi-hop reasoning. As shown in \figref{fig:visualmodel}, Track Module aims to enrich the representation of the question from the visual information by exploiting the question and the dialog history. Locate Module aims to enrich the representation of the question from the textual dialog history by exploiting the question and the image. Answer decoder takes the outputs of Track Module and Locate module as inputs, and generates an appropriate response.

We first introduce the representations of inputs (both image features and the language features). Then we describe the detailed architectures of dual-channel multi-hop reasoning and multimodal fusion operation. Finally, we present the multimodal attention answer decoder.

\begin{figure}[t!]
\centering
\scalebox{1}{
  \begin{overpic}[width=\columnwidth]{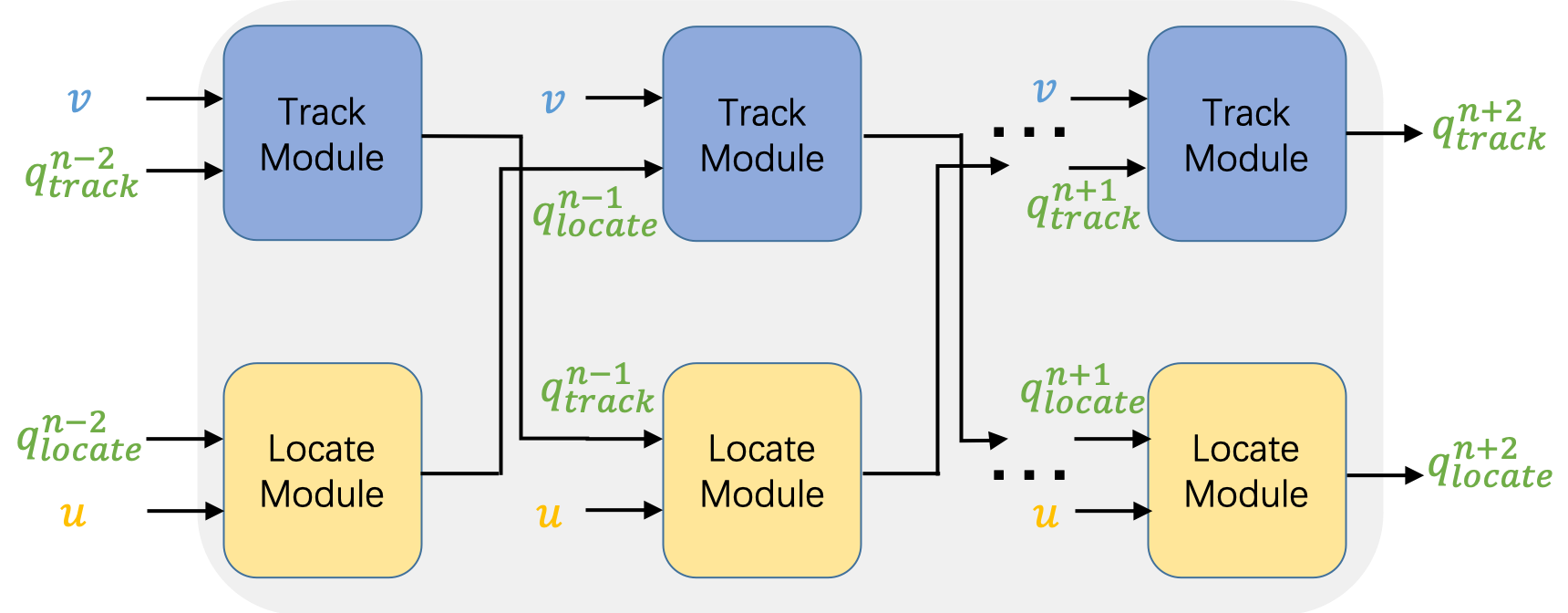}
   \end{overpic}
   }
   \caption{Schematic representation of multi-hop reasoning. Please see Section Dual-channel Multi-hop Reasoning for details. All $q$ at different hops denote different query features, $v$ denotes the image features and $u$ denotes the dialog history features.
   }\label{fig:submodel}

\end{figure}

\subsection{Input Representation}\label{section:input}
\paragraph{Image Features} 
We use a pre-trained Faster R-CNN~\cite{ren2015faster} to extract object-level image features. Specifically, the image features $v$ for the image $I$ are represented by:
\begin{equation}
  v = {\rm Faster\ R-CNN}(I) \in \mathbb{R}^{K \times V} \label{v1},
\end{equation} 
where $K$ denotes the total number of the object detection features per image and $V$ denotes the dimension of each features, respectively. We extract the object features by using a fixed number $K$.

\paragraph{Language Features}
We first embed each word in the current question $Q_t$ to $\{w_{t,1}, \dots, w_{t,L}\}$ by using pre-trained Glove embeddings~\cite{pennington2014glove}, where $L$ denotes the number of tokens in $Q_t$. We use a one-layer BiLSTM to generate a sequence of hidden states $\{x_{t,1}, \dots, x_{t,L}\}$. We use the last hidden state of the BiLSTM as question features $q_t \in \mathbb{R}^L$ as follows:
\begin{eqnarray}
  \overrightarrow{x_{t,j}} &=& {\rm LSTM}_f(w_{t,j}, x_{t,j-1}),\ j\in \{0, \dots, L-1\},\\
  \overleftarrow{x_{t,j}} &=& {\rm LSTM}_b(w_{t,j}, x_{t,j+1}),\ j\in \{L-1, \dots, 0\},\\
  q_t &=& [\overrightarrow{x_{t,L-1}}, \overleftarrow{x_{t,0}}], \label{eq:u1}
\end{eqnarray}

Also, each question-answer pair in the dialog history $H = \{H_0, H_1, \dots, H_{t-1}\}$ and the answer $A_t$ are embedded in the same way as the current question, yielding the dialog history features $u = \{u_0, u_1, \dots, u_{t-1}\}$ and the answer features $a_t$. $Q_t$, $H$ and $A_t$ are embedded with the same word embedding vectors but three different BiLSTMs.

\subsection{Dual-channel Multi-hop Reasoning}\label{section:dualchannel} 

The dual-channel multi-hop reasoning framework is implemented via two modules, i.e., Track Module and Locate Module. Track Module aims to make a fully understanding of the question from the aspect of the image. Locate Module aims to make a fully understanding of the question from the aspect of the dialog history. The multi-hop reasoning pathway of Track Module is illustrated as $I_1 \to H_2 \to I_3 \cdots \to I_n$ and the multi-hop reasoning pathway of Locate Module is illustrated as $H_1 \to I_2 \to H_3 \cdots \to H_n$. Next, we formally describe the single-hop Track Module and Locate Module, and then extend them to multi-hop ones. We use the 3-hop reasoning in this paper.

\paragraph{Track Module} 
Track Module is designed to help enrich the semantic presentation of the question from the image. In order to obtain the question- and history-aware representation of the image, we implement Track Module by taking the inspiration from bottom-up attention mechanism~\cite{anderson2018bottom}. Track Module takes the query features $q_{track}$ (for instance, the question feature $q$ at reasoning hop 1) and image features $v$ (\equref{v1}) as inputs, and then outputs query-aware representation of the image. We first project these two vectors to $d_{track}$ dimension and compute soft attention over all the object detection features as follows:
\begin{eqnarray}
  S &=& f^q_{track}(q_{track}) \circ f^v_{track}(v), \label{eq:track1}\\
  \alpha &=& {\rm softmax}(W^S S +b^S),\label{eq:track2}
\end{eqnarray} 
where $f_{track}^q(\cdot)$ and $f_{track}^v(\cdot)$ denote the 2-layer perceptrons with ReLU activation which transform the dimension of input features to $d_{track}$, $W^S$ is the project matrix for the softmax activation and $\circ$ denotes Hadamard product. From these equations, we get the query-aware attention weights $\alpha \in \mathbb{R}^{K \times 1}$. Next we apply the query-aware attention weights to image features $v$ to compute the query-aware representation of the image as follows:
\begin{eqnarray}
  q^{out}_{track} = \sum_{j=1}^K\alpha_{j}v_{j} \label{eq:track3}.
\end{eqnarray}
We use $\rm{Track}(\cdot, \cdot)$ to represent the operations of Track Module, namely~\equref{eq:track1} -~\equref{eq:track3}, here and after.

Furthermore, we use Track Module in the multi-hop reasoning process to enrich the semantic presentation of the question from the image. Details are to be formalized in Section Multi-hop Reasoning.

\paragraph{Locate Module}
Locate Module is designed to get a rich representation of the question from the dialog history. 
Similar with Track Module, Locate Module takes the query features $q_{locate}$ (for instance, the question feature $q$ at reasoning hop 1) and dialog history features $u$ (\equref{eq:u1}) as inputs, and then outputs query-aware representation of the dialog history features as follows:
\begin{eqnarray}
  Z &=& f^q_{locate}(q_{locate}) \circ f^u_{locate}(u), \label{eq:locate1}\\
  \eta &=& {\rm softmax}(W^Z Z +b^Z),\label{eq:locate2}
\end{eqnarray} 
where $f_{locate}^q(\cdot)$ and $f_{locate}^v(\cdot)$ denote the two layer multi-layer perceptrons with ReLU activation which transform the dimension of input features to $d_{locate}$, $W^Z$ is the project matrix for the softmax activation and $\circ$ denotes Hadamard product. From these equations, we get the query-aware attention weights $\eta \in \mathbb{R}^{T \times 1}$. Next we apply the query-aware attention weights to the dialog history features $u$ to compute the query-aware representation of the dialog history as follows:
\begin{eqnarray}
  \hat{u} = \sum_{j=1}^T\eta_{j}u_{j} \label{eq:locate3}.
\end{eqnarray}
Next we apply $\hat{u}$ to two layer multi-layer perceptrons with ReLU activation in between, then add it with the representation of the caption $u_0$. Layer normalization~\cite{kang2019dual} is also applied in this step.
\begin{eqnarray}
  g &=& W^2_u{\rm ReLU}(W^1_u\hat{u}+b^1_u)+b^2_u,\\
  q^{out}_{locate} &=& {\rm LayerNorm}(g + u_0)\label{eq:locate5}.
\end{eqnarray} 
We use $\rm{Locate}(\cdot, \cdot)$ to represent the operations of locate module, namely~\equref{eq:locate1} -~\equref{eq:locate5}, here and after.

Furthermore, we use Locate Module in the multi-hop reasoning process to enrich the semantic presentation of the question from the dialog history. Details are to be formalized in Section Multi-hop Reasoning.

\paragraph{Multi-hop Reasoning}\label{section:multihop}
Dual-channel multi-hop reasoning contains two types of multi-hop reasoning. One is multi-hop reasoning, starting from and ending with the image, illustrated as $I_1 \to H_2 \to I_3 \cdots \to I_n$. The other one is multi-hop reasoning, starting from and ending with the dialog history, illustrated as $H_1 \to I_2 \to H_3 \cdots \to H_n$. We implement each reasoning pathway via Track Module and Locate Module. The reasoning pathway $I_1 \to H_2 \to I_3 \cdots \to I_n$ includes the following steps: 

\begin{eqnarray*}
  {\rm step}\ 1: & {\rm Track}(q,v) \to q_{track}^1; \\
  {\rm step}\ 2:& {\rm Locate}(q_{track}^1, u) \to q_{track}^2;\\
  {\rm step}\ 3:& {\rm Track}(q_{track}^2,v) \to q_{track}^3;\\
  &\cdots\\
  {\rm step}\ n:& {\rm Track}(q_{track}^{n-1},v) \to q_{track}^n.
\end{eqnarray*}
The reasoning pathway $H_1 \to I_2 \to H_3 \cdots \to H_n$ includes the following steps:
\begin{eqnarray*}
  {\rm step}\ 1:& {\rm Locate}(q,u) \to q_{locate}^1; \\
  {\rm step}\ 2:& {\rm Track}(q_{locate}^1, v) \to q_{locate}^2;\\
  {\rm step}\ 3:& {\rm Locate}(q_{locate}^2, u) \to q_{locate}^3;\\
  &\cdots\\
  {\rm step}\ n:& {\rm Locate}(q_{locate}^{n-1}, u) \to q_{locate}^n.
\end{eqnarray*}
Parameters of modules at each reasoning hop are not shared with the other. Note that the reasoning process is valid only if $n$ is an odd number. In this paper, we use 3-hop reasoning for Visual Dialog.

\subsection{Multimodal Fusion}\label{section:mfdecoder}
In this section, we introduce multimoal fusion. As shown in~\figref{fig:visualmodel}, before we fuse the multimodal representations $q_{track}^n$ and $q_{locate}^n$ generated by Track Module and Locate Module, we use question features $q$ to enhance the representations $q_{track}^n$ and $q_{locate}^n$ as follows:
\begin{eqnarray}
  \hat{q}_{track}^n &=& f^q_{att}(q) \circ f^h_{att}(q_{track}^n), \label{eq:attenhance1}\\
  \hat{q}_{locate}^n &=& f^q_{att}(q) \circ f^v_{att}(q_{locate}^n)\label{eq:attenhance2},
\end{eqnarray}
where $f^q_{att}(\cdot)$, $f^h_{att}(\cdot)$ and $f^v_{att}(\cdot)$ denote 2-layer perceptrons with ReLU activation. Both~\equref{eq:attenhance1} and~\equref{eq:attenhance2} are named as the Att-Enhance module. We also use Att-Enhance modules between 2-hop and 3-hop. Then we fuse the representations of two channels as follows:
\begin{eqnarray}
  e &=& [W^1_f\hat{q}_{track}^n + b^1_f,  W^2_f\hat{q}_{locate}^n + b^2_f,],\\
  \hat{e} &=& {\rm tanh}(W^3_fe + b^3_f) , \label{eq:e}
\end{eqnarray}
where $[\cdot]$ denotes the concatenation operation, $W^1_f$, $W^2_f$, $W^3_f$ and $b^1_f$, $b^2_f$, $b^2_f$ are learned parameters.

\begin{figure}[t!]
\centering
\scalebox{0.8}{
  \begin{overpic}[width=\columnwidth]{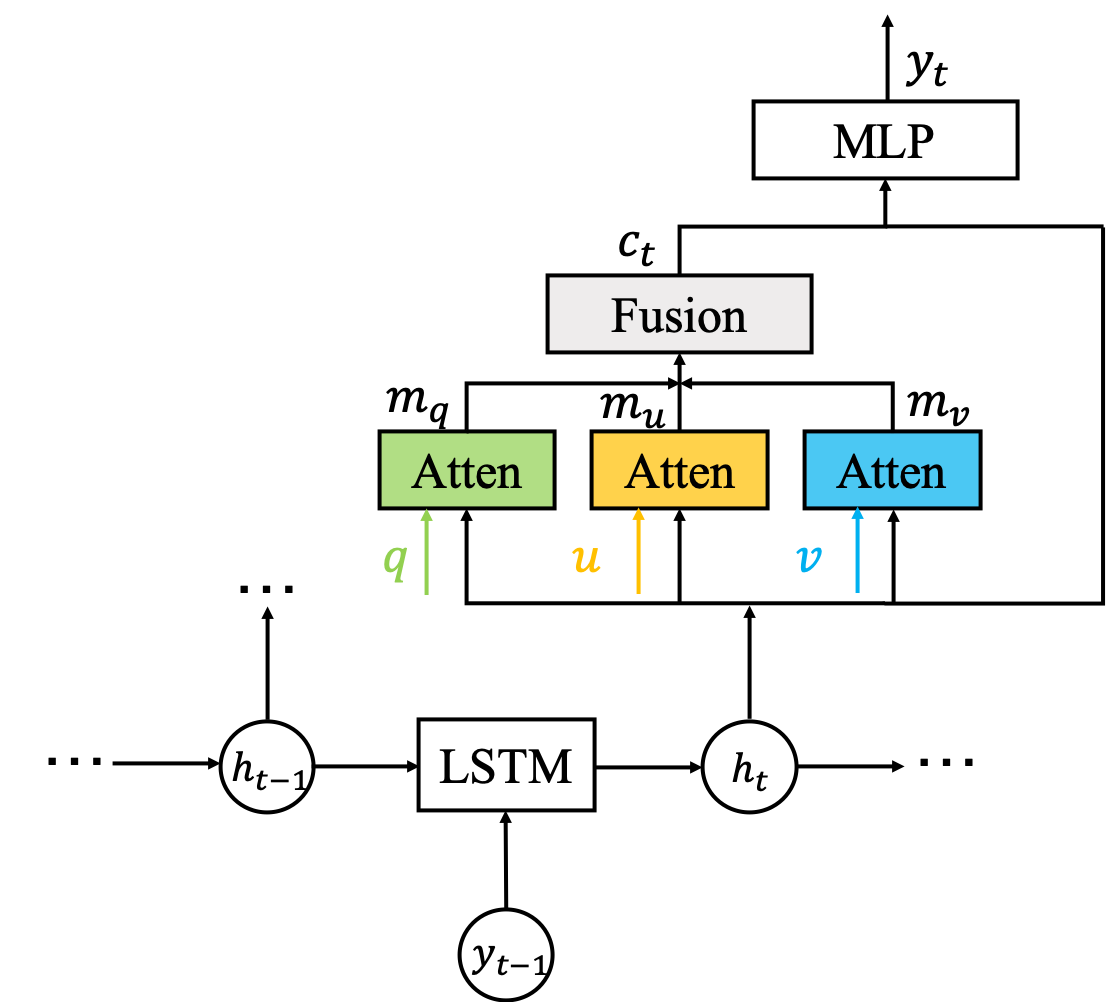}
  \end{overpic}
  }
  \caption{Multi-modal Attention Decoder. We use the multi-modal context vector $e$ (\equref{eq:e}) to initial the deocder LSTM, utilize hidden $h_t$ to attend to the question features $q$, history features $u$, image features $v$ and combine the attended representations $m_q$, $m_u$, $m_v$ to predict the next word united with hidden $h_t$. 
  }\label{fig:decoder}
\end{figure}

\subsection{Generative Decoder}
\label{section:decoder}
As illustrated in \figref{fig:decoder}, our generative decoder is adapted from spatial attention based decoders~\cite{lu2017knowing}. In the encoder-decoder framework, with recurrent neural network (RNN), we model the conditional probability as:
\begin{equation}
    p(y_t|y_1,\dots, y_{t-1}, q, v, u) = f(h_t, c_t),
\end{equation}
where $f$ is 2-layer perceptrons with ReLU activation, $c_t$ is the mulitmodal context vector at time $t$ and $h_t$ is the hidden state of the RNN at time $t$. In this paper, we use LSTM and $h_t$ is modeled as:
\begin{equation}
    h_t = {\rm LSTM}(y_{t-1}, h_{t-1}),
\end{equation}
where $y_{t-1}$ is the representation of the generative answer at time step $t-1$.

Given the question features $q$, dialog history features $u$, image features $v$, and hidden state $h_t$, we feed them through a 1-layer perceptron with a softmax function to generate the three attention distribution over the question, $T$ rounds of the history and $K$ object detection features per image, respectively. First, the attended question vector $m_q$ is as defined:
\begin{eqnarray}
    z^q_t &=& {W^q_h}{\rm tanh}(W_qq+(W^q_gh_t)\mathbbm{1}^T), \label{eq:w1}\\
    \alpha^q_t &=& {\rm softmax}(z^q_t),
\end{eqnarray}
where $\mathbbm{1}$ is a vector with all elements set to 1, $W_q$, $W^q_g$, $W^q_h$ are learned parameters. All the bias terms in description of \equref{eq:w1} and \equref{eq:w2} are omitted for simplicity. Then we obtain the attended question vector $m_q$ as follows:
\begin{equation}
m_q = \sum_{i=1}^l\alpha^q_{t,i}q_i.
\end{equation}
Similar with the computation of attended question, we obtain the attended history vector $m_u$ and attended image vector $m_v$. Then we fuse these three context vectors to obtain the context vector $c_t$ by:
\begin{equation}
    c_t = {\rm tanh}(W_c[m_q, m_h, m_v]), \label{eq:w2}
\end{equation}
where $[\cdot]$ denotes concatenation and $w_c$ is learned parameters. $c_t$ and $h_t$ are combined to predict next word $y_{t+1}$.

In addition, we use the encoder output $\hat{e}$ as embedding input to initialize our decoder LSTM. Formally,
\begin{equation}
    h_0 = {\rm LSTM}(\hat{e}, s_q),
\end{equation}
where $s_q$ is the last state of the question LSTM in the encoder and $h_0$ is used as the initial state of the decoder LSTM. 

\section{Experiments}\label{section:exp}

\subsection{Datasets} We evaluate our proposed approach on the VisDial v0.9 and v1.0 datasets~\cite{das2017visual}. VisDial v0.9 contains 83k dialog on COCO-train~\cite{lu2017best} and 40k dialog on COCO-val~\cite{lu2017best} images, for a total of 1.23M dialog question-answer pairs. VisDial v1.0 dateset is an extension of VisDial v0.9 dateset with an additional 10k COCO-like images from Flickr. Overall, VisDial v1.0 dateset contains 123k (all images from v0.9), 2k and 8k images as train, validation and test splits, respectively.

\begin{table}
  \centering
    \resizebox{1\columnwidth}!{
  \begin{tabular}{lccccc}
    \toprule
    Model & MRR & R@1 & R@5 & R@10 & Mean\\
    \midrule
    AP~\cite{das2017visual}& 37.35 & 23.55 & 48.52 & 53.23 & 26.50 \\
    NN~\cite{das2017visual} & 42.74 & 33.13 & 50.83 & 58.69 & 19.62 \\
    LF~\cite{das2017visual} & 51.99 & 41.83 & 61.78 & 67.59 & 17.07 \\
    HREA~\cite{das2017visual} & 52.42 & 42.28 & 62.33 & 68.71 & 16.79 \\
    MN~\cite{das2017visual} & 52.59 & 42.29 & 62.85 & 68.88 & 17.06 \\
    HCIAE~\cite{lu2017best} & 53.86 & 44.06 & 63.55 & 69.24 & 16.01 \\
    CoAtt~\cite{wu2018you} & 54.11 & 44.32 & 63.82 & 69.75 & 16.47 \\
    CoAtt~\cite{wu2018you}$^\dagger$ & 55.78 & 46.10 & 65.69 & 71.74 & 14.43 \\
    RvA~\cite{niu2019recursive} & 55.43 & 45.37 & 65.27 & \bf{72.97} & \bf{10.71} \\
    \midrule
    DMRM & \bf{55.96} & \bf{46.20} & \bf{66.02} & 72.43 & 13.15\\
    \bottomrule
  \end{tabular}
}
  \caption{Performance on VisDial val v0.9~\cite{das2017visual}. Higher the better for mean reciprocal rank (MRR) and recall@$k$ (R@1, R@5, R@10), while lower the better for mean rank. Our proposed model outperforms all other models on MRR, R@5, and mean rank. $\dagger$ indicates that the model is trained by using reinforcement learning. 
  }\label{tab:resultv0.9}
  \end{table}
  
\begin{table}
  \centering
    \resizebox{1\columnwidth}!{
  \begin{tabular}{lccccc}
    \toprule
    Model & MRR & R@1 & R@5 & R@10 & Mean\\
    \midrule
    MN~\cite{das2017visual}$^\ddagger$ & 47.99 & 38.18 & 57.54 & 64.32 & 18.60 \\
    HCIAE~\cite{lu2017best}$^\ddagger$ & 49.10 & 39.35 & 58.49 & 64.70 & 18.46 \\
    CoAtt~\cite{wu2018you}$^\ddagger$ & 49.25 & 39.66 & 58.83 & 65.38 & 18.15 \\
    ReDAN~\cite{gan2019multi} & 49.69 & \bf{40.19} & 59.35 & 66.06 & 17.92 \\
    \midrule
    DMRM & \bf{50.16} & 40.15 & \bf{60.02} & \bf{67.21} & \bf{15.19} \\
    \bottomrule
  \end{tabular}
 }
  \caption{Performance on VisDial val v1.0~\cite{das2017visual}. $\ddagger$ All the models are re-implemented by \citeauthor{gan2019multi}~\shortcite{gan2019multi}.
  }\label{tab:resultv1.0}
  \end{table}

\subsection{Evaluation Metrics} 
We follow~\citeauthor{das2017visual}~\shortcite{das2017visual} to use a retrieval setting to evaluate the individual responses at each round of a dialog. Specifically, at test time, apart from the image, ground truth dialog history and the question, a list of 100 candidates answers are also given. The model is evaluated on retrieval metrics: (1) rank of human response, (2) existence of the human response in $top-k$ ranked responses, i.e., recall@$k$ and (3) mean reciprocal rank (MRR) of the human response. Since we focus on evaluating the generalization ability of our generator, the sum of the log-likelihood of each option is used for ranking.

\subsection{Implementation Details} To process the data, we first lowercase all the texts, convert digits to words, then remove contractions before tokenizing. The captions, questions and answers are further truncated to ensure that they are no longer than 24, 16 or 8 tokens, respectively. We then construct the vocabulary of tokens that appear at least 5 times in the training split, giving us a vocabulary of 8,958 words on VisDial v0.9 and 10,366 words on VisDial v1.0. 
All the BiLSTMs in our model are 1-layered with 512 hidden states. The Adam optimizer~\cite{kingma2014adam} is used with the base learning rate of 1e-3, further decreasing to 1e-5 with a warm-up process.

\subsection{Results and Analysis} We compare our proposed models to the state-of-the-art generative models developed in previous works. As shown in ~\tabref{tab:resultv0.9} and ~\tabref{tab:resultv1.0}, our proposed model achieves the state-of-the-art results on some metrics on the VisDial v0.9 and v1.0 datasets. The key observations are as follows:
\begin{itemize}
  \item By comparing with singe-hop approaches (LF~\cite{das2017visual} and HCAIE~\cite{lu2017best}), we demonstrate the validity of multi-hop reasoning, because it utilizes the abundant latent information between modalities. 
  \item By comparing with singe-channel approaches (CoAtt~\cite{wu2018you} and RvA~\cite{niu2019recursive}), we come to the conclusion that dual-channel reasoning is beneficial for gaining an original understanding of the question from the dialog history and the image.
  \item By comparing with other methods and the state-of-the-art approaches (HREA~\cite{das2017visual}, MN~\cite{das2017visual} and ReDAN~\cite{gan2019multi}), our approach achieves the state-of-the-art results on some metrics that demonstrate the superiority of our model.
\end{itemize}

\paragraph{Ablation Study}As shown in~\tabref{tab:ablationstudy}, different settings explain the importance of each part of our model. By comparing ``DMRM w/ n-hop'', we see the effectiveness of multi-hop reasoning. By comparing ``DMRM w/o Locate'' and ``DMRM w/o Track'' with DMRM, we see the effectiveness of our dual-channel models. By comparing our final model DMRM with ``DMRM w/o AttD'', we illustrates the improvement due to multimodal attention in the decoder.

\begin{table}
  \centering
    \resizebox{1\columnwidth}!{
  \begin{tabular}{lccccc}
    \toprule
    Model & MRR & R@1 & R@5 & R@10 & Mean\\
    \midrule
    DMRM w/ 1-hop & 55.04 & 45.55& 64.46 & 70.49 & 14.68\\
    DMRM w/ 2-hop & 54.87 & 44.85 & 65.05 & 71.75 & 13.66\\
    DMRM w/ 3-hop & 55.57 & 45.80 & 65.54 & 72.09 & 13.51\\
    \midrule
    DMRM w/o Locate & 54.77 & 45.35 & 64.04 & 70.01 & 14.81 \\
    DMRM w/o Track & 53.28 & 43.06 & 63.47 & 70.06 & 14.54 \\
    DMRM w/o AttD & 55.57 & 45.80 & 65.54 & 72.09 & 13.51 \\
    \midrule
    DMRM & \bf{55.96} & \bf{46.20} & \bf{66.02} & \bf{72.43} & \bf{13.15} \\
    \bottomrule
  \end{tabular}
  }
  \caption{Ablation study of our proposed model on VisDial val v0.9~\cite{das2017visual}. ``DMRM w/ n-hop'' means the model use n-hop reasoning. 
  ``DMRM w/o AttD'' means the model is not use multimodal attention decoder. Note that ``DMRM w/ 2-hop'' is an incomplete reasoning process under our designed architecture and the ablation study of n-hop reasoning is based on the model ``DMRM w/o AttD''.}\label{tab:ablationstudy}
  \end{table}

\paragraph{Significance Test} We use t-test and analysis of variance (ANOVA) to analyze results of sentences generated by our model and the HCIAE model~\cite{lu2017best}. The p-values of these two analytical methods are all less than 0.01, indicating that the results are significantly different.

\begin{table}
  \centering
    \resizebox{0.95\columnwidth}!{
  \begin{tabular}{p{5cm}cc}
    \toprule
      & HCIAE & Ours\\
    \midrule
    Human evaluation method 1 (M1): & 0.60 & \bf{0.65}\\
    \midrule
    Human evaluation method 2 (M2): & 0.53 & \bf{0.62}\\
    \bottomrule
  \end{tabular}
  }
  \caption{Human evaluation on 100 sampled responses on VisDial val v0.9.  M1: percentage of responses pass the Turing Test. M2: percentage of responses evaluated as better or equal to human responses. }\label{tab:sigtest}
\end{table}

\paragraph{Human Evaluation} We randomly extract 100 samples for human evaluation. The evaluation results are as shown in~\tabref{tab:sigtest}, which show the effectiveness of our model.

\begin{figure*}[t!]
    \centering
    \scalebox{0.94}{
  \begin{overpic}[width=0.94\textwidth]{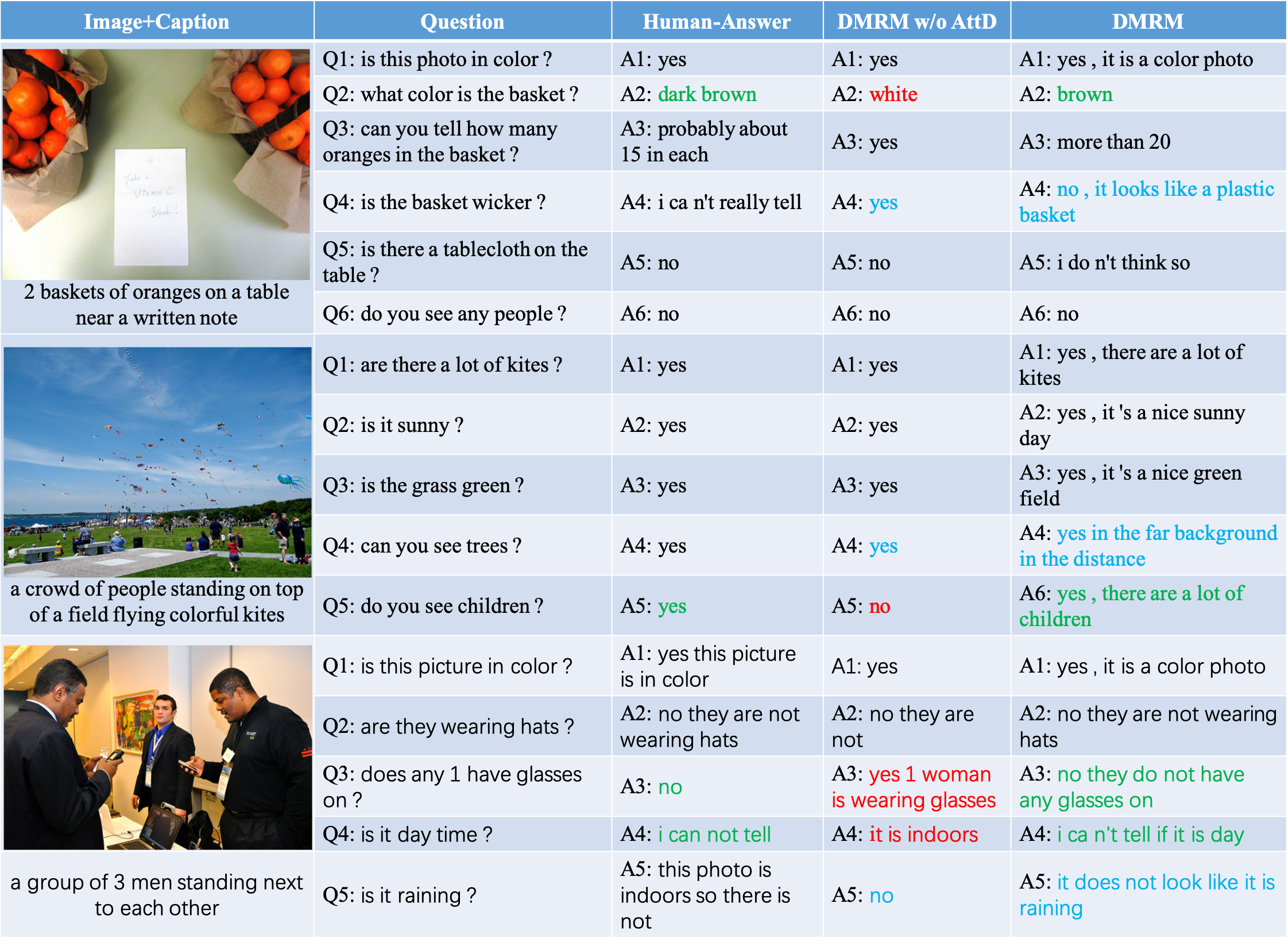}
   \end{overpic}
   }
   \caption{Qualitative results of our final model (DMRM) on VisDial v0.9 comparing to human ground-truth answers and our baseline model (``DMRM w/o AttD''). Compared with ``DMRM w/o AttD'', DMRM utilizes the muliti-modal attention in the decdoer. The improvement of correctness (marked in green and red) and interpretability (marked in blue) of generated answers due to our multi-modal attention in the decoder are partially colored.
   }\label{fig:examples}
\end{figure*}

\begin{figure*}[t!]
\centering
\scalebox{0.94}{
  \begin{overpic}[width=0.94\textwidth]{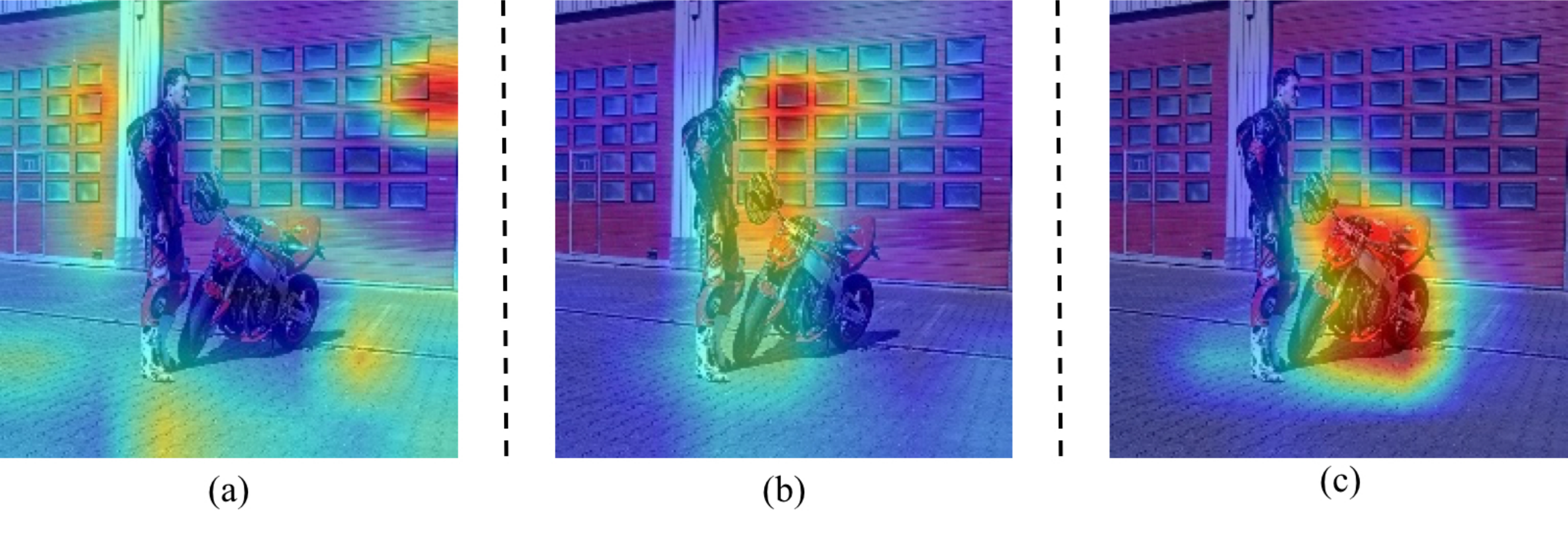}
   \end{overpic}
   }
   \caption{Visualization of our reasoning process. (a) The attended image at 1-hop via Track Module. (b) The attended image at 2-hop via Track Module. (c) The attended image at 3-hop via Track Module. With the question ``what color is his bike ?'', our model finally attends to the bike to get the answer. 
   }\label{fig:attention}
\end{figure*}

\paragraph{Quantitative Results Analysis} As shown in \figref{fig:examples}, our model generate responses of a high degree of consistency with human answers, which shows the effectiveness of it. Compared with ``DMRM w/o AttD'', DMRM generate more correct and meaningful responses.
\figref{fig:attention} is the visualization of our reasoning process. For the question ``{\em what color is his bike ?}'', the model infers step by step, finally pays attention to the bike to answer the current question.

\section{Related Work}

\paragraph{Vision-language Task}
Vision-language tasks, such as image caption~\cite{ren2015exploring,gao2015you,kinghorn2018region,tan2019phrase,ding2019image} and visual question answering (VQA)~\cite{yang2016stacked,anderson2018bottom,alberti2019fusion,cadene2019murel,vedantam2019probabilistic}, have aroused great interest in recent years. Image caption is a task of describing the visual content of an image by using one or more sentences while visual question answering focuses on providing a natural language answer given an image and free-form, open question. Visual dialog~\cite{wu2018you,lu2017best,seo2017visual,guo2019image} can be seen as an extension of image caption and VQA tasks. Visual dialog enables an AI agent not only to interact with the visual environment but also to have a continuous conversation with human.

\paragraph{Visual Dialog}
Visual dialog has attracted widespread attention. Some previous works are similar to our work, but fundamentally different from ours.
\citeauthor{das2017visual}~\shortcite{das2017visual} propose a dialog-RNN, which takes the question, the image and the last round history as inputs, and then produces both an encoding representation for this round and a dialog context for the next round. \citeauthor{das2017visual}~\shortcite{das2017visual} exploit a dialog-RNN to deal with the mutil-turn dialog only by using the information of last round history while we leverage a multi-hop reasoning for visual dialog at each turn and exploit the whole dialog history.
Besides,~\citeauthor{gan2019multi}~\shortcite{gan2019multi} provide a multi-step reasoning model via a RNN, which firstly leverage the query and history to attend to the image, secondly use the query and the image to attend to the history, and finally utilize the image and the history to update RNN State. Nevertheless, we propose the dual-channel multi-hop reasoning via two modules where Track Module only deals with the image and Locate Module only utilizes the information of the dialog history. Moreover, we conduct the representation of the question to the representation generated by Track Module and Locate module between reasoning hops.

\section{Conclusion}

We introduce our Dual-channel Multi-hop Reasoning Model (DMRM) for visual dialog, a new framework to simultaneously capture information from the dialog history and the image to enrich the semantic representation of the question by exploiting dual-channel reasoning. 
This dual-channel mulit-hop reasoning process provides a more fine-grained understanding of the question by utilizing the textual information and the visual context simultaneously via multi-hop reasoning, thus boosting the performance of answer generation. Experiments conducted on the VisDial v0.9 and v1.0 certify the effectiveness of our proposed method.

\section{Acknowledgments}
This work was supported by the Major Project for New Generation of AI (Grant No. 2018AAA0100400), the National Natural Science Foundation of China (Grant No. 61602479), and the Strategic Priority Research Program of the Chinese Academy of Sciences (Grant No. XDB32070000).

\bibliography{aaai-2020}

\begin{thebibliography}{}

\bibitem[\protect\citeauthoryear{Alberti \bgroup et al\mbox.\egroup
  }{2019}]{alberti2019fusion}
Alberti, C.; Ling, J.; Collins, M.; and Reitter, D.
\newblock 2019.
\newblock Fusion of detected objects in text for visual question answering.
\newblock In {\em Proceedings of the 2019 Conference on Empirical Methods in
  Natural Language Processing and the 9th International Joint Conference on
  Natural Language Processing},  2131--2140.

\bibitem[\protect\citeauthoryear{Anderson \bgroup et al\mbox.\egroup
  }{2016}]{Anderson2016SPICE}
Anderson, P.; Fernando, B.; Johnson, M.; and Gould, S.
\newblock 2016.
\newblock {SPICE}: Semantic propositional image caption evaluation.
\newblock {\em Adaptive Behavior} 11(4):382--398.

\bibitem[\protect\citeauthoryear{Anderson \bgroup et al\mbox.\egroup
  }{2018}]{anderson2018bottom}
Anderson, P.; He, X.; Buehler, C.; Teney, D.; Johnson, M.; Gould, S.; and
  Zhang, L.
\newblock 2018.
\newblock Bottom-up and top-down attention for image captioning and visual
  question answering.
\newblock In {\em Proceedings of the IEEE Conference on Computer Vision and
  Pattern Recognition},  6077--6086.

\bibitem[\protect\citeauthoryear{Cadene \bgroup et al\mbox.\egroup
  }{2019}]{cadene2019murel}
Cadene, R.; Ben-Younes, H.; Cord, M.; and Thome, N.
\newblock 2019.
\newblock Murel: Multimodal relational reasoning for visual question answering.
\newblock In {\em Proceedings of the IEEE Conference on Computer Vision and
  Pattern Recognition},  1989--1998.

\bibitem[\protect\citeauthoryear{Das \bgroup et al\mbox.\egroup
  }{2017}]{das2017visual}
Das, A.; Kottur, S.; Gupta, K.; Singh, A.; Yadav, D.; Moura, J.~M.; Parikh, D.;
  and Batra, D.
\newblock 2017.
\newblock Visual dialog.
\newblock In {\em Proceedings of the IEEE Conference on Computer Vision and
  Pattern Recognition},  326--335.

\bibitem[\protect\citeauthoryear{Ding \bgroup et al\mbox.\egroup
  }{2019}]{ding2019image}
Ding, S.; Qu, S.; Xi, Y.; Sangaiah, A.~K.; and Wan, S.
\newblock 2019.
\newblock Image caption generation with high-level image features.
\newblock {\em Pattern Recognition Letters} 123:89--95.

\bibitem[\protect\citeauthoryear{Gan \bgroup et al\mbox.\egroup
  }{2019}]{gan2019multi}
Gan, Z.; Cheng, Y.; Kholy, A.~E.; Li, L.; Liu, J.; and Gao, J.
\newblock 2019.
\newblock Multi-step reasoning via recurrent dual attention for visual dialog.
\newblock In {\em Proceedings of the 57th Annual Meeting of the Association for
  Computational Linguistics},  6463--6474.

\bibitem[\protect\citeauthoryear{Gao \bgroup et al\mbox.\egroup
  }{2015}]{gao2015you}
Gao, H.; Mao, J.; Zhou, J.; Huang, Z.; Wang, L.; and Xu, W.
\newblock 2015.
\newblock Are you talking to a machine? dataset and methods for multilingual
  image question.
\newblock In {\em Advances in Neural Information Processing Systems},
  2296--2304.

\bibitem[\protect\citeauthoryear{Guo, Xu, and Tao}{2019}]{guo2019image}
Guo, D.; Xu, C.; and Tao, D.
\newblock 2019.
\newblock Image-question-answer synergistic network for visual dialog.
\newblock In {\em Proceedings of the IEEE Conference on Computer Vision and
  Pattern Recognition},  10434--10443.

\bibitem[\protect\citeauthoryear{Hu \bgroup et al\mbox.\egroup
  }{2018}]{hu2018explainable}
Hu, R.; Andreas, J.; Darrell, T.; and Saenko, K.
\newblock 2018.
\newblock Explainable neural computation via stack neural module networks.
\newblock In {\em Proceedings of the European Conference on Computer Vision},
  53--69.

\bibitem[\protect\citeauthoryear{Hudson and
  Manning}{2018}]{hudson2018compositional}
Hudson, D.~A., and Manning, C.~D.
\newblock 2018.
\newblock Compositional attention networks for machine reasoning.
\newblock In {\em International Conference on Learning Representations}.

\bibitem[\protect\citeauthoryear{Kang, Lim, and Zhang}{2019}]{kang2019dual}
Kang, G.-C.; Lim, J.; and Zhang, B.-T.
\newblock 2019.
\newblock Dual attention networks for visual reference resolution in visual
  dialog.
\newblock In {\em Proceedings of the 2019 Conference on Empirical Methods in
  Natural Language Processing and the 9th International Joint Conference on
  Natural Language Processing},  2024--2033.

\bibitem[\protect\citeauthoryear{Kinghorn, Zhang, and
  Shao}{2018}]{kinghorn2018region}
Kinghorn, P.; Zhang, L.; and Shao, L.
\newblock 2018.
\newblock A region-based image caption generator with refined descriptions.
\newblock {\em Neurocomputing} 272:416--424.

\bibitem[\protect\citeauthoryear{Kingma and Ba}{2014}]{kingma2014adam}
Kingma, D.~P., and Ba, J.
\newblock 2014.
\newblock Adam: A method for stochastic optimization.
\newblock {\em arXiv preprint arXiv:1412.6980}.

\bibitem[\protect\citeauthoryear{Lu \bgroup et al\mbox.\egroup
  }{2016}]{lu2016hierarchical}
Lu, J.; Yang, J.; Batra, D.; and Parikh, D.
\newblock 2016.
\newblock Hierarchical question-image co-attention for visual question
  answering.
\newblock In {\em Advances In Neural Information Processing Systems},
  289--297.

\bibitem[\protect\citeauthoryear{Lu \bgroup et al\mbox.\egroup
  }{2017a}]{lu2017best}
Lu, J.; Kannan, A.; Yang, J.; Parikh, D.; and Batra, D.
\newblock 2017a.
\newblock Best of both worlds: Transferring knowledge from discriminative
  learning to a generative visual dialog model.
\newblock In {\em Advances in Neural Information Processing Systems},
  314--324.

\bibitem[\protect\citeauthoryear{Lu \bgroup et al\mbox.\egroup
  }{2017b}]{lu2017knowing}
Lu, J.; Xiong, C.; Parikh, D.; and Socher, R.
\newblock 2017b.
\newblock Knowing when to look: Adaptive attention via a visual sentinel for
  image captioning.
\newblock In {\em Proceedings of the IEEE Conference on Computer Vision and
  Pattern Recognition},  375--383.

\bibitem[\protect\citeauthoryear{Niu \bgroup et al\mbox.\egroup
  }{2019}]{niu2019recursive}
Niu, Y.; Zhang, H.; Zhang, M.; Zhang, J.; Lu, Z.; and Wen, J.-R.
\newblock 2019.
\newblock Recursive visual attention in visual dialog.
\newblock In {\em Proceedings of the IEEE Conference on Computer Vision and
  Pattern Recognition},  6679--6688.

\bibitem[\protect\citeauthoryear{Pennington, Socher, and
  Manning}{2014}]{pennington2014glove}
Pennington, J.; Socher, R.; and Manning, C.
\newblock 2014.
\newblock Glove: Global vectors for word representation.
\newblock In {\em Proceedings of the 2014 Conference on Empirical Methods in
  Natural Language Processing},  1532--1543.

\bibitem[\protect\citeauthoryear{Ren \bgroup et al\mbox.\egroup
  }{2015}]{ren2015faster}
Ren, S.; He, K.; Girshick, R.; and Sun, J.
\newblock 2015.
\newblock Faster r-cnn: Towards real-time object detection with region proposal
  networks.
\newblock In {\em Advances in Neural Information Processing Systems},  91--99.

\bibitem[\protect\citeauthoryear{Ren, Kiros, and
  Zemel}{2015}]{ren2015exploring}
Ren, M.; Kiros, R.; and Zemel, R.
\newblock 2015.
\newblock Exploring models and data for image question answering.
\newblock In {\em Advances in Neural Information Processing Systems},
  2953--2961.

\bibitem[\protect\citeauthoryear{Seo \bgroup et al\mbox.\egroup
  }{2017}]{seo2017visual}
Seo, P.~H.; Lehrmann, A.; Han, B.; and Sigal, L.
\newblock 2017.
\newblock Visual reference resolution using attention memory for visual dialog.
\newblock In {\em Advances in Neural Information Processing Systems},
  3719--3729.

\bibitem[\protect\citeauthoryear{Tan and Chan}{2019}]{tan2019phrase}
Tan, Y.~H., and Chan, C.~S.
\newblock 2019.
\newblock Phrase-based image caption generator with hierarchical lstm network.
\newblock {\em Neurocomputing} 333:86--100.

\bibitem[\protect\citeauthoryear{Vedantam \bgroup et al\mbox.\egroup
  }{2019}]{vedantam2019probabilistic}
Vedantam, R.; Desai, K.; Lee, S.; Rohrbach, M.; Batra, D.; and Parikh, D.
\newblock 2019.
\newblock Probabilistic neural symbolic models for interpretable visual
  question answering.
\newblock In {\em Proceedings of International Conference on Machine Learning},
   6428--6437.

\bibitem[\protect\citeauthoryear{Wu \bgroup et al\mbox.\egroup
  }{2018}]{wu2018you}
Wu, Q.; Wang, P.; Shen, C.; Reid, I.; and van~den Hengel, A.
\newblock 2018.
\newblock Are you talking to me? reasoned visual dialog generation through
  adversarial learning.
\newblock In {\em Proceedings of the IEEE Conference on Computer Vision and
  Pattern Recognition},  6106--6115.

\bibitem[\protect\citeauthoryear{Xu \bgroup et al\mbox.\egroup
  }{2015}]{xu2015show}
Xu, K.; Ba, J.; Kiros, R.; Cho, K.; Courville, A.; Salakhudinov, R.; Zemel, R.;
  and Bengio, Y.
\newblock 2015.
\newblock Show, attend and tell: Neural image caption generation with visual
  attention.
\newblock In {\em Proceedings of International Conference on Machine Learning},
   2048--2057.

\bibitem[\protect\citeauthoryear{Yang \bgroup et al\mbox.\egroup
  }{2016}]{yang2016stacked}
Yang, Z.; He, X.; Gao, J.; Deng, L.; and Smola, A.
\newblock 2016.
\newblock Stacked attention networks for image question answering.
\newblock In {\em Proceedings of the IEEE Conference on Computer Vision and
  Pattern Recognition},  21--29.

\end{thebibliography}
\bibliographystyle{aaai}

\end{document}